\documentclass[11pt]{article}
\usepackage{graphicx}
\usepackage{amstext}
\usepackage{amsmath}

\usepackage{url}

\newcommand{\VAR}[1]{\text{Var}\left[#1 \right]}

\newcommand\independent{\protect\mathpalette{\protect\independenT}{\perp}}
\def\independenT#1#2{\mathrel{\rlap{$#1#2$}\mkern2mu{#1#2}}}

\newcommand{\NIL}[1]{}
\newcommand{\ee}{\mathbf e}

\newcommand{\m}{\mathbf m}
\newcommand{\cc}{\mathbf c}
\newcommand{\sss}{\mathbf s}

\newcommand{\Z}{\mathbf Z}
\newcommand{\z}{\mathbf z}

\newcommand{\w}{\mathbf w}

\newcommand{\s}{\mathbf s}

\begin{document}

\title{From dependency to causality: a machine learning approach}

\author{Gianluca Bontempi, Maxime Flauder\\
       Machine Learning Group, Computer Science Department,\\
       Interuniversity Institute of Bioinformatics in Brussels (IB)$^2$,\\
        ULB, Universit\'{e} Libre de Bruxelles,\\
       Brussels, Belgium\\
       email: {\tt gbonte@ulb.ac.be} \\
      }

\maketitle

\begin{abstract}
The relationship between statistical dependency and causality lies at the heart
of all statistical approaches to causal inference. Recent results in the ChaLearn cause-effect pair challenge have shown that causal directionality
can be inferred with good accuracy also in Markov indistinguishable configurations thanks to
data driven approaches.
This paper proposes a supervised machine learning approach to infer the existence
of a directed causal link between two variables in multivariate settings with  $n>2$ variables. The approach relies on the asymmetry
of some conditional (in)dependence relations between the members of the Markov blankets  of 
two variables causally connected.
Our results show that supervised learning methods may be successfully
used to extract causal information on the basis of asymmetric statistical descriptors 
also for $n>2$ variate distributions.

\end{abstract}

\section{Introduction}
\label{sec:introduction}
The relationship between statistical dependency and causality lies at the heart
of all statistical approaches to causal inference and can be summarized by two famous
statements: \emph{correlation (or more generally statistical association) does not imply causation} and \emph{causation induces 
a statistical dependency between causes and effects (or more generally descendants)}~(\cite{Reichenbach56}).
In other terms it is well known that statistical dependency is a necessary yet not sufficient condition for causality.
The unidirectional link between these two notions has been used by many formal approaches to causality
to justify the adoption of statistical methods for detecting or inferring causal links from observational data.
The most influential one is the Causal Bayesian Network approach, detailed in~(\cite{koller09friedman}) which relies 
on notions of independence and conditional independence to detect causal patterns in the data.
Well known examples of related inference algorithms are the constraint-based methods like the PC algorithms~(\cite{Spirtes2000}) and IC~(\cite{pearl00}). 
These approaches are  founded on probability theory and have been shown to be accurate in reconstructing causal patterns in many 
 applications. At the same time they restrict the set of configurations which causal inference is applicable to. Such boundary is essentially determined
by the notion of \emph{distinguishability} which defines the set of Markov equivalent configurations on the basis of conditional independence tests.
Typical examples of indistinguishability are the two-variable setting and the completely connected triplet configuration~(\cite{Guyon07AliferisElisseeff})
where it is impossible to distinguish between cause and effects by means of conditional 
or unconditional independence tests.

If on one hand the notion of indistinguishability is probabilistically sound, on the other hand it contributed
to slow down the development of alternative methods
to address interesting yet indistinguishable causal patterns.
The slow down was in our opinion due to a misunderstanding of the meaning 
and the role of the notion of indistinguishability. Indistiguishability results rely on two main
aspects: i) they refer only to specific features of dependency (notably conditional or unconditional independence)
and ii) they state the conditions (e.g. faithfulness) under which it is possible to distinguish (or not) \emph{with certainty} between configurations.

Accordingly, indistinguishability results do not prevent the existence of statistical
algorithms able to \emph{reduce the uncertainty about the causal pattern} even in indistinguishable 
configurations.
This has been made evident by the appearance in recent years of a series of approaches which tackle
the cause-effect pair inference, like ANM (Additive Noise Model)~(\cite{Hoyer09etal}), IGCI (Information Geometry Causality Inference)~(\cite{Daniusis10etal,Janzing12etal}), LiNGAM (Linear Non Gaussian Acyclic Model)~(\cite{LINGAM})  and
the algorithms described in~(\cite{Mooij10etal}) and~(\cite{statnikov12etal})\footnote{
A more extended list of recent algorithms is available in http://www.causality.inf.ethz.ch/cause-effect.php?page=help}.  What is common to these approaches is
that they use alternative statistical features of the data to detect causal patterns and reduce the 
uncertainty about their directionality. A further important step  in this direction has been
represented by the recent organization of the ChaLearn cause-effect pair challenge~(\cite{guyon14}).
The  good (and significantly better than random) accuracy obtained  on the basis of observations
of pairs of causally related (or unrelated) variables supports the idea
that alternative strategies can be designed to infer with success (or at least
significantly better than random) indistinguishable configurations.

It is worthy to remark that the best ranked approaches\footnote{We took part in the ChaLearn challenge and we ranked 8th in the final leader board.} in the ChaLearn competition share a common aspect: they  infer from statistical features 
of the bivariate distribution the probability of the existence and 
then of the directionality of the causal link between two variables.
The success of these approaches shows that the problem of causal inference
can be successfully addressed  as a supervised machine learning approach where the inputs
are features describing the probabilistic dependency and the output
is a class denoting the existence (or not) of a directed causal link. Once sufficient training data are made
available, conventional feature selection algorithms~(\cite{guyon03}) and classifiers can be used
to return a prediction.

The effectiveness of machine learning strategies in the case of pairs of variables
encourages  the extension of the strategy to configurations with a larger number of variables.
In this paper we propose an original approach to learn from multivariate observations 
the probability that a variable is a direct cause of another. This task is undeniably
more difficult because  
\begin{itemize}
\item  the number of parameters needed to describe a multivariate distribution increases rapidly (e.g. quadratically
in the Gaussian case),
\item    information about the existence of a causal link between two variables is returned also by the nature of the dependencies existing between the two  variables and the remaining ones.
\end{itemize}
The second consideration is evident in the case of a collider configuration $\z_1 \rightarrow \z_2 \leftarrow \z_3$: in this case
the dependency (or independency) between $\z_1$ and $\z_3$ tells us more about the link $\z_1 \rightarrow \z_2$
than the dependency between $\z_1$ and $\z_2$.
This led us to develop a machine learning strategy (described in Section~\ref{sec:more}) where descriptors of the relation
existing between members of the Markov blankets of two variables are used to learn the probability (i.e. a score) that a causal 
link exists between two variables. 
The approach relies on the asymmetry of some conditional (in)dependence relations between the members of the Markov blankets of two variables causally connected. The resulting algorithm (called D2C and described in Section~\ref{sec:D2C}) predicts the existence of a direct causal link
between two variables in a multivariate setting by (i) creating a set of of features of
the relationship based on asymmetric descriptors of the multivariate dependency and (ii) using a  classifier
to learn a mapping between the features and the presence of a causal link.

In Section~\ref{sec:expe} we report the results of a set of experiments assessing the accuracy
of the D2C algorithm. Experimental results based on synthetic and published data show that the D2C approach is competitive
and often outperforms state-of-the-art methods.

\section{Learning the relation between dependency and causality in a configuration with $n>2$ variables.}
\label{sec:more}

This section presents an approach to learn, from 
a number of observations, the relationships existing between the $n$ variate
distribution of $\Z=[\z_1,\dots,\z_n]$ and the existence of a directed causal link between  
two variables $\z_i$ and $\z_j$, $1 \le i \neq j \le n$. 
Several parameters may be estimated from data in order to represent
the multivariate distribution of $\Z$, like the correlation or the partial correlation matrix. 
Two problems however arise in this case: (i) these
parameters are informative in case of Gaussian distributions only and (ii) identical (or close)
causal configurations could be associated to very different parametric values, thus making difficult the learning of
the mapping.

In other terms it is more relevant to describe the distribution in structural
terms (e.g. with notions of conditional dependence/independence) rather than in parametric terms. 
Two more aspects have to be taken into consideration.
First since we want to use a learning approach to identify cause-effect relationships
we need some quantitative features to describe the structure of the multivariate distribution.
Second, since asymmetry is a distinguishing characteristic of a causal relationship, we expect
that effective features should share the same asymmetric properties. 

In this paper we will use information theory to
represent and quantify the notions of (conditional) dependence and independence
between variables and to derive a set of asymmetric features  to reconstruct  causality from dependency.

\subsection{Notions of information theory}
Let us consider three continuous random variables $\z_1$, $\z_2$ and $\z_3$
having a joint Lebesgue
density\footnote{Boldface
denotes random variables.}.
Let us start by considering the relation between $\z_1$
and $\z_2$.
The mutual information~(\cite{cover90}) between $\z_1$ and $\z_2$
is defined in terms of their probabilistic density functions
$p(z_1)$, $p(z_2)$ and $p(z_1,z_2)$ as
\begin{equation}
\label{eq:mi}
I(\z_1;\z_2)=\int \int \log \frac{p(z_1,z_2)}{p(z_1)p(z_2)} p(z_1,z_2) dz_1 dz_2
=H(\z_1)-H(\z_1| \z_2)
\end{equation}
where $H$
is the \emph{entropy} and  the convention $0 \log \frac{0}{0}=0$ is adopted.
This quantity measures the amount of stochastic dependence
between $\z_1$ and $\z_2$~(\cite{cover90}).
Note that, if $\z_1$ and $\z_2$ are Gaussian distributed the following relation holds
\begin{equation}
\label{eq:corr}
I(\z_1;\z_2)=-\frac{1}{2} \log (1-\rho^2)
\end{equation}
where $\rho$ is the Pearson correlation coefficient between $\z_1$ and $\z_2$. 

Let us now consider a third variable $\z_3$. The \emph{conditional mutual 
information}~(\cite{cover90}) between $\z_1$ and $\z_2$ once $\z_3$ is given
is defined by
\begin{multline}
\label{eq:cond_mi}
I(\z_1;\z_2|\z_3)=\int \int \int \log \frac{p(z_1,z_2|z_3)}{p(z_1|z_3)p(z_2|z_3)} p(z_1,z_2,z_3) dz_1 dz_2 dz_3=\\
=H(\z_1| \z_3)-H(\z_1| \z_2,\z_3)
\end{multline}
The conditional  mutual information is null if and only if $\z_1$ and $\z_2$
are conditionally independent given $\z_3$.

A structural notion which can described in terms of conditional mutual
information is the notion of Markov Blanket (MB). The Markov Blanket of variable $\z_i$ in an $n$ dimensional
distribution is the smallest subset of variables belonging to $\Z \setminus \z_i$ (where $\setminus$ denotes the set difference operator) which makes $\z_i$ conditionally independent of all the remaining ones.
In information theoretic terms let us consider a set $\Z$ of $n$ random variables, a  variable $\z_i$ and a subset ${\mathbf M_i}
 \subset \Z \setminus \z_i$.  The subset ${\mathbf M_i}$ is said to be 
a \emph{Markov blanket} of $\z_i$  if it is the minimal subset satisfying 
\begin{equation*}
I(\z_i; (\Z \setminus {(\mathbf M_i \cup \z_i)}) |{\mathbf M_i})=0
\end{equation*}

Effective algorithms have been proposed in literature to infer a Markov Blanket from
observed data~(\cite{tsamardinos03aliferis}). Feature selection algorithms are also
useful to construct a Markov blanket of a given target variable once 
they rely on notions of conditional independence to select relevant variables~(\cite{meyer2014bontempi}).

\subsection{Causality and asymmetric dependency relationships}
The notion of causality is central in science and also an intuitive notion of everyday life. 
The remarkable property of causality which distinguishes it from dependency  is asymmetry.

In probabilistic terms a variable $\z_i$ is dependent on a variable $\z_j$ if the density of $\z_i$, conditional on the observation $\z_j=z_j$, is different from the marginal one    
$$
p(z_i| \z_j=z_j) \neq p(z_i)
$$
In information theoretic terms the two variables are dependent if $I(\z_i;\z_j)=I(\z_j;\z_i) >0$. This implies that dependency is \emph{symmetric}. If $\z_i$ is dependent on $\z_j$, then $\z_j$ is dependent on $\z_i$ too as shown by 
$$
p(z_j| \z_i=z_i) \neq p(z_j)
$$

The formal representation of the notion of causality demands an extension of the syntax of the
probability calculus as done by~\cite{Pearl95}  with the introduction of the operator {\tt  do}
which allows to distinguish the observation of a value of $\z_j$ (denoted by $\z_j=z_j$) from the manipulation of the variable $\z_j$ (denoted by $\tt{ do} (\z_j=z_j)$). 
Once this extension is accepted we say that a variable $\z_j$ is a cause of a variable $\z_i$ (e.g. "diseases cause symptoms") if the distribution of $\z_i$ is different from the marginal one when we set the value $\z_j=z_j$  
$$
p(z_i | {\tt  do} (\z_j=z_j)) \neq p(z_i)
$$
but not viceversa (e.g. "symptoms do not cause disease")
$$
p(z_j| {\tt  do} (\z_i=z_i)) = p(z_j)
$$
The extension of the probability notation made by Pearl allows to formalize the intuition that causality is \emph{asymmetric}.
Another notation which allows to represent causal expression is provided by graphical models or more
specifically by Directed Acyclic Graphs (DAG)~(\cite{koller09friedman}). 
In this paper we will limit to consider causal relationships modeled by DAG, 
which proved to be convenient tools to understand and use the notion of causality.
Furthermore we will make the assumption that the set of causal relationships
existing between the variables of interest can be described by a Markov and faithful DAG~(\cite{pearl00}).
This means that the DAG is an accurate map of dependencies and independencies of the represented distribution
and that using the notion of \emph{d-separation} it is possible to read from the graph if two sets of nodes are (in)dependent conditioned on a third.

\begin{figure}
\vskip 0.2in
\begin{center}
\includegraphics[width=\columnwidth]{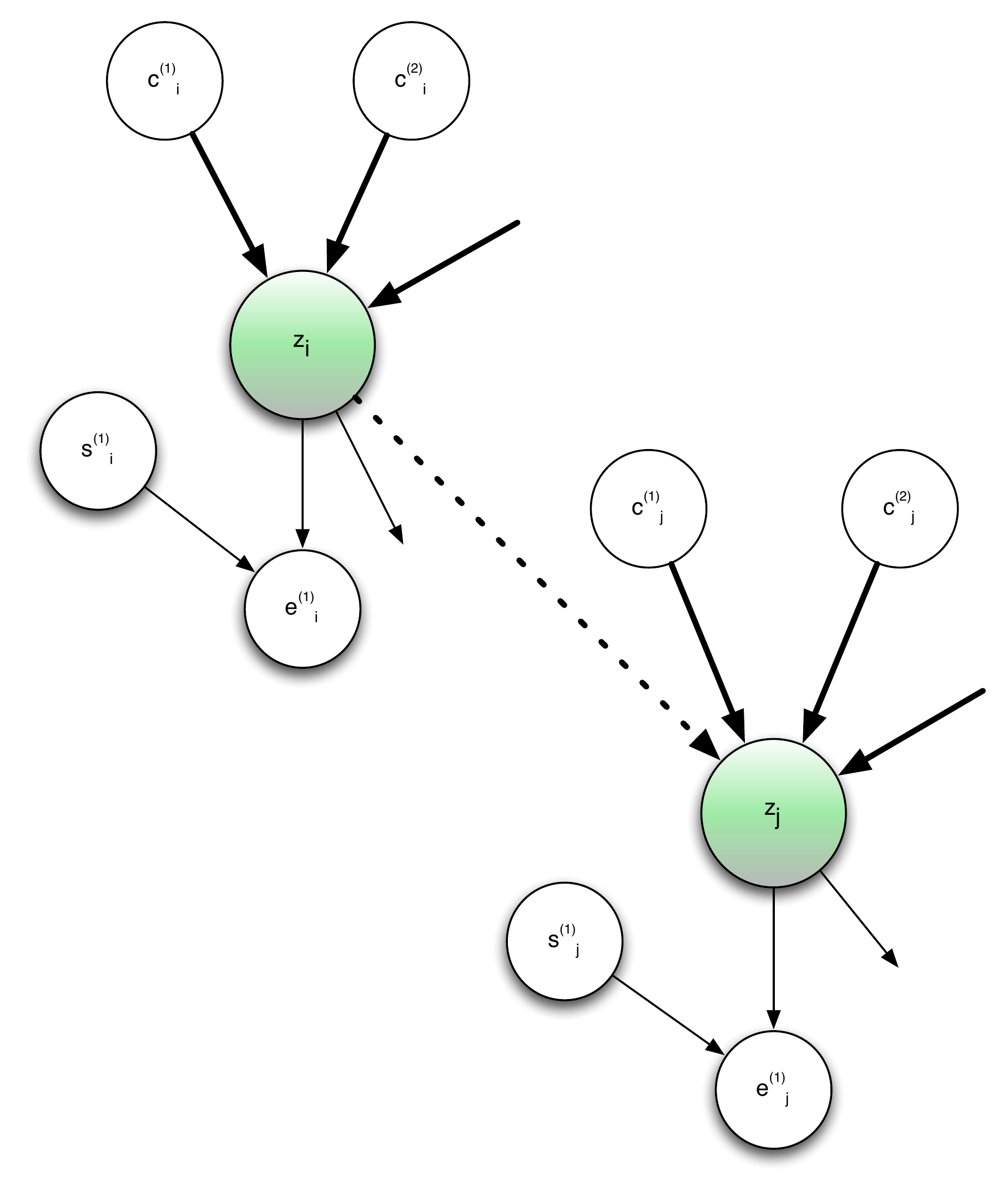}
\caption{Two causally connected variables and their Markov Blankets. \label{fig:MB}}
\end{center}
\vskip -0.2in
\end{figure}

The asymmetric nature of causality suggests that if we want to infer causal links from dependency we need to find 
some features (or descriptors)  which describe the dependency and share with causality the  property of asymmetry.
Let us suppose that we are interested in predicting the existence of a directed causal link $\z_i \rightarrow \z_j$ 
where $\z_i$ and $\z_j$ are components of  an observed $n$-dimensional vector $\Z=[\z_1,\dots,\z_n]$.

We define as \emph{dependency descriptor}  of the ordered pair $\langle i,j \rangle$ a function $d(i,j)$ of the distribution of $\Z$
which depends on $i$ and $j$.
Example of dependency descriptors are the correlation $\rho(i,j)$ between $\z_i$ and $\z_j$, the mutual information $I(\z_i;\z_j)$ or 
the partial correlation between $\z_i$ and $\z_j$ given another variable
$\z_k, i \neq j \neq k$.

We define a dependency descriptor \emph{symmetric} if $d(i,j)=d(j,i)$ otherwise we call it \emph{asymmetric}.
Correlation or mutual information are symmetric descriptors since
$$
d(i,j)=I(\z_i;\z_j)=I(\z_j;\z_i)=d(j,i)
$$

Because of the asymmetric property of causality,  if we want to maximize our chances to reconstruct
causality from dependency we need to identify relevant asymmetric descriptors.
In order to define useful asymmetric descriptors we have recourse to the Markov
Blankets of the two variables $\z_i$ and $\z_j$.

Let us consider for instance the portion of a DAG represented in Figure~\ref{fig:MB} where the variable $\z_i$
is a direct cause of $\z_j$. The figure shows also the Markov Blankets of the two
variables (denoted $M_i$ and $M_j$ respectively)  and their components, i.e. the direct causes (denoted by $\cc$), the direct effects ($\ee$)
and the spouses ($\sss$)~(\cite{Pellet08Elisseeff}). 

In what follows we will make two assumptions: (i) the only  connection between the two sets is the edge $\z_i \rightarrow \z_j$ and (ii) there is no common ancestor of $\z_i$ ($\z_j$) and its spouses $\sss_i$ ($\sss_j$).
We will discuss  these assumptions at the end of the section.
Given these assumptions and because of d-separation,  a number of  asymmetric conditional (in)dependence relations 
holds between the members of of $M_i$ and $M_j$ (Table~\ref{tab:sep}). 
For instance (first line of Table~\ref{tab:sep}), by conditioning on the  effect $\z_j$ we create a dependence between $\z_i$ and the direct causes of $\z_j$  
while by conditioning on the
 $\z_i$ we d-separate $\z_j$ and the direct causes of $\z_i$.

The relations in Table~\ref{tab:sep} can be used to define the following set of asymmetric descriptors, 
\begin{align}
\label{eq:D2}
& d^{(k)}_1(i,j)=I(\z_i;\cc^{(k)}_j | \z_j),\\
\label{eq:D3}
& d^{(k)}_2(i,j)=I(\ee^{(k)}_i;\cc^{(k)}_j | \z_j),\\
\label{eq:D4}
& d^{(k)}_3(i,j)=I(\cc^{(k)}_i;\cc^{(k)}_j | \z_j),\\
\label{eq:D1}
& d^{(k)}_4(i,j)=I(\z_i;\cc^{(k)}_j),
\end{align}
whose asymmetry  is given by 
 \begin{align}
 \label{eq:aD2}
 & d^{(k)}_1(i,j)=I(\z_i;\cc^{(k)}_j | \z_j) >0, \quad d^{(k)}_1(j,i)=I(\z_j;\cc^{(k)}_i | \z_i) =0,\\
 \label{eq:aD3}
 & d^{(k)}_2(i,j)=I(\ee^{(k)}_i;\cc^{(k)}_j | \z_j) >0, \quad d^{(k)}_2(j,i)=I(\ee^{(k)}_j;\cc^{(k)}_i | \z_i)=0, \\
 \label{eq:aD4}
 & d^{(k)}_3(i,j)=I(\cc^{(k)}_i;\cc^{(k)}_j | \z_j) >0, \quad d^{(k)}_3(j,i)=I(\cc^{(k)}_j;\cc^{(k)}_i | \z_i)=0,\\
  \label{eq:aD1}
 & d^{(k)}_4(i,j)=I(\z_i;\cc^{(k)}_j)=0, \quad d^{(k)}_4(j,i)=I(\z_j;\cc^{(k)}_i)>0. 
 \end{align}

\begin{table}
\center
\begin{tabular}{ c||c}
Relation $i,j$  &  Relation $j,i$\\
\hline
 $\forall k \quad \z_i  \not \independent \cc^{(k)}_j | \z_j $  & $\forall k \quad \z_j   \independent \cc^{(k)}_i | \z_i $ \\
  $\forall k \quad \ee^{(k)}_i \not \independent \cc^{(k)}_j | \z_j $ & $ \forall k \quad \ee^{(k)}_j  \independent \cc^{(k)}_i | \z_i $\\
$\forall k \quad \cc^{(k)}_i \not \independent \cc^{(k)}_j | \z_j $ & $ \forall k \quad \cc^{(k)}_j  \independent \cc^{(k)}_i | \z_i $\\
 $\forall k \quad \z_i  \independent \cc^{(k)}_j $  & $\forall k \quad   \z_j    \not \independent  \cc^{(k)}_i $\\
\hline
%
\hline
\end{tabular}
\caption{Asymmetric (un)conditional (in)dependance relationships between members of the Markov Blankets of $\z_i$ and $\z_j$ in Figure~\ref{fig:MB}. \label{tab:sep}}
\end{table}

\begin{table}
\center
\begin{tabular}{ c||c}
Relation $i,j$  &  Relation $j,i$\\
\hline
 $\forall k \quad \z_i  \not \independent \ee^{(k)}_j $  & $\forall k \quad   \z_j  \not \independent \ee^{(k)}_i$ \\
$\forall k \quad \z_i   \independent \s^{(k)}_j $  & $\forall k \quad   \z_j   \independent \s^{(k)}_i$ \\
$\forall k \quad \z_i   \independent \ee^{(k)}_j | \z_j $  & $\forall k \quad  \z_j  \independent \ee^{(k)}_i | \z_i$ \\
$\forall k \quad \z_i   \independent \s^{(k)}_j | \z_j $  & $\forall k \quad  \z_j  \independent \s^{(k)}_i | \z_i $ \\
$\forall k \quad \ee^{(k)}_i   \independent \ee^{(k)}_j | \z_i $  & $\forall k \quad  \ee^{(k)}_j   \independent \ee^{(k)}_i | \z_j$ \\
$\forall k \quad \ee^{(k)}_i   \independent \s^{(k)}_j | \z_j $  & $\forall k \quad  \ee^{(k)}_j   \independent \s^{(k)}_i | \z_i$ \\
\hline

\hline
\end{tabular}
\caption{Symmetric (un)conditional (in)dependance relationships between members of the Markov Blankets of $\z_i$ and $\z_j$ in Figure~\ref{fig:MB}. \label{tab:sep2}}
\end{table}

At the same time we can write a set of symmetric conditional (in)dependence relations (Table~\ref{tab:sep2}) and the equivalent formulations in terms of mutual information terms:
 \begin{align}
 \label{eq:sD1}
& I(\z_j;\ee^{(k)}_i)>0,\\
\label{eq:sD2}
& I(\z_i;\ee^{(k)}_j)>0,\\
  \label{eq:sD3}
& I(\z_j;\sss^{(k)}_i)=I(\z_i;\sss^{(k)}_j)=0,\\
  \label{eq:sD4}
& I(\z_i;\ee_j^{(k)}|\z_j)= I(\z_j;\ee_i^{(k)}|\z_i) =I(\z_i;\sss_j^{(k)}|\z_j)= I(\z_j;\sss_i^{(k)}|\z_i)=0,\\
  \label{eq:sD5}
& I(\ee_i^{(k)};\ee_j^{(k)}|\z_j)= I(\ee_j^{(k)};\ee_i^{(k)}|\z_i) =I(\ee_i^{(k)};\sss_j^{(k)}|\z_j)=I(\ee_j^{(k)};\sss_i^{(k)}|\z_i) =0.
\end{align}

\subsection{From asymmetric relationships to distinct distributions}
\label{sec:as_distr}
The asymmetric properties of the four descriptors (\ref{eq:D2})-(\ref{eq:D1}) is encouraging if we want to exploit 
dependency related features to infer causal properties from data.
However, this optimism is  undermined by the fact that all the descriptors 
require already the capability of distinguishing between the causes (i.e. the terms $\cc$) and the effects (i.e. the terms $\ee$)  of the Markov Blanket 
of a given variable. Unfortunately this discriminating capability is what we are looking for!

In order to escape this circularity problem we consider two solutions. The first is to have recourse
to a preliminary phase that prioritizes the components of the Markov Blanket and then use this result as starting point
to detect asymmetries and then improve the classification of causal links. This is for instance feasible by  using a filter selection algorithm, like mIMR~(\cite{bontempi2010causal}),
which aims to prioritize the direct causes in the Markov Blanket by searching for pairs of variables with high relevance
and low interaction.

The second solution is related to the fact that the asymmetry of the four descriptors  induces
a difference in the distributions of some information theoretic terms which do not require the distinction between causes and effects within the Markov Blanket. The 
consequence is that we can replace the descriptors~(\ref{eq:D2})-(\ref{eq:D1}) with other 
descriptors (denoted with the letter $D$) that can be actually estimated from data.

Let $\m^{(k)}$ denote  a generic component of the Markov Blanket with no distinction between cause, effect or spouse.
It follows that a population made of terms depending on $\m^{(k)}$ is a mixture of three subpopulations,
the first made of causes, the second made of effects and the third of spouses, respectively.
It follows that the distribution of the population is a \emph{finite mixture}~(\cite{McLaughlan2000}) of three
distributions, the first related to the causes, the second to the effects and the third to the spouses.
Since the moments of the finite mixture are functions of the moments of each component, we can derive 
some properties of the resulting mixture from the properties of each component. For instance
if we can show that two subpopulations are identical but that all the elements of the third subpopulation in the first mixture are larger than the elements of the analogous subpopulation in the second mixture, we can derive that the two mixture distributions are different.

Consider for instance the quantity $I(\z_i;\m^{(k_j)}_j | \z_j)$ 
where $\m^{(k_j)}_j$, $k_j=1,\dots,K_j$ is a member of the set $M_j \setminus \z_i$. 
From~(\ref{eq:aD2}) and~(\ref{eq:sD4}) it follows that the  mixture distribution associated to the populations ${D}_1(i,j)=\{I(\z_i;\m_j^{(k_j)}|\z_j), k_j=1,\dots,K_j\}$ and ${D}_1(j,i)=\{ I(\z_j;\m_i^{(k_i)}|\z_i), k_i=1,\dots,K_i\}$ 
are different since 
\begin{equation}
\begin{cases}
I(\z_i;\m_j^{(k_j)}|\z_j)> I(\z_j;\m_i^{(k_i)}|\z_i), \quad & \mbox{ if } \m_j^{(k_j)}=\cc_j^{(k_j)} \land \m_i^{(k_i)}=\cc_i^{(k_i)}\\
I(\z_i;\m_j^{(k_j)}|\z_j)=I(\z_j;\m_i^{(k_i)}|\z_i), \quad & \mbox{else}
\end{cases}
\end{equation}

It follows that even if we are not able to distinguish between a cause $\cc_j \in M_j$ and an effect $\ee_j\in M_j$,
we know that the distribution of  the population ${D}_1(i,j)$ differs from the distribution of the population ${D}_1(j,i)$.
We can therefore use the population ${D}_1(i,j)$ (or some of its moments) as descriptor of the causal dependency.

Similarly we can replace the descriptors~(\ref{eq:D3}),~(\ref{eq:D4}) with the distributions of the population ${D}_2(i,j)=\{ I(\m_i^{(k_i)};\m_j^{(k_j)}|\z_j) , k_j=1,\dots,K_j , k_i=1,\dots,K_i \}$.
From~(\ref{eq:aD3}),~(\ref{eq:aD4}) and~(\ref{eq:sD5}) we obtain that 
the distributions of the populations ${D}_2(i,j)$ and ${D}_2(j,i)$ are different.

If we make the additional assumption that $I(\z_j;\ee^{(k)}_i)= I(\z_i;\ee^{(k)}_j)>0$ from~(\ref{eq:aD1})  we obtain also 
 that the distribution of the population ${D}_3(i,j)= \{ I(\z_i;\m_j^{(k_j)}), k_j=1,\dots,K_j \}$ is different from the one of  ${D}_3(j,i)=\{I(\z_j;\m_i^{(k_i)})$, $k_i=1,\dots,K_i\}$.

The previous results are encouraging and show that though we are not able to distinguish between 
the different components of a Markov Blanket, we can notwithstanding compute 
some quantities (in this case distributions of populations) whose asymmetry is informative about the causal relationships $\z_i \rightarrow \z_j$.

As a consequence by measuring from observed data some statistics (e.g. quantiles) related to the distribution 
of these asymmetric descriptors, we may obtain some insight about the causal relationship between two variables.
This idea is made explicit in the algorithm described in the following section.

Though these results rely on the two assumptions made before, two considerations  are worthy to be made. First, the 
main goal of the approach  is to shed light on the existence of dependency asymmetries also in multivariate contributions.   Secondly  we expect that the second layer (based on supervised learning) will eventually compensate for configurations not compliant with the assumptions and  take advantage of complementarity or synergy of the  descriptors in discriminating between causal configurations.

\section{The D2C algorithm}
\label{sec:D2C}
The rationale of the D2C algorithm is to predict the existence of a causal link
between two variables in a multivariate setting by (i) creating a set of features of
the relationship between the members of the Markov Blankets of the two variables and (ii) using a  classifier
to learn a mapping between the features and the presence of a causal link.

We use two sets of features to summarize the relation between the two Markov blankets: the first one accounts for the presence (or the position if the MB is obtained by ranking)  
of the terms of $M_j$ in $M_i$
and viceversa.  For instance it is evident that if $\z_i$ is a cause of $\z_j$ we expect to find $\z_i$ highly ranked between
the causal terms of $M_j$ but $\z_j$  absent (or ranked low) among the causes of $M_i$.
The second set of features is based on the results of the previous section and is obtained by summarizing the
distributions of the asymmetric descriptors with a set of quantiles.

We propose then an algorithm (D2C) which for each pair of measured variables $\z_i$ and $\z_j$:
\begin{enumerate}
\item infers from  data the two Markov Blankets (e.g. by using state-of-the-art approaches) $M_i$
and $M_j$ and the subsets
$M_i \setminus \z_j = \{ \m^{(k_i)}, k_i=1,\dots,K_i\}$ and $M_j \setminus \z_i= \{ \m^{(k_j)}, k_j=1,\dots,K_j \}$.
Most of the existing algorithms associate to the Markov Blanket a ranking such that the most strongly relevant variables
are ranked before.
\item computes a set of (conditional) mutual information terms describing the dependency between $\z_i$ and $\z_j$
$$I=[I(\z_i; \z_j), I(\z_i; \z_j | \mbox{M}_j \setminus \z_i) , I(\z_i; \z_j | \mbox{M}_i \setminus \z_j) ] $$

\item computes the positions $P_i^{(k_i)}$ of the members $\m^{(k_i)}$ of $M_i\setminus \z_j$ in the ranking associated to $M_j \setminus \z_i$ and
the positions $P_j^{(k_j)}$ of the  terms  $\m^{(k_j)}$ in the ranking associated to $M_i\setminus \z_j$. Note that in case of the absence of a term of $M_i$ in $M_j$, the position is set to $K_j+1$ (respectively $K_i+1$).
\item computes the populations based on the asymmetric descriptors introduced in Section~\ref{sec:as_distr}:
\begin{enumerate}
\item $ {D}^{(1)}(i,j)= \{ I(\z_i;\m^{(k_j)}_j | \z_j), k_j=1,\dots,K_j\} $ 
\item $ {D}^{(1)}(j,i)=\{ I(\z_j;\m^{(k_i)}_i | \z_i), k_i=1,\dots,K_i\}  $
\item $ {D}^{(2)}(i,j)=\{I(\m^{(k_i)}_i;\m^{(k_j)}_j | \z_i), k_i=1,\dots,K_i,  k_j=1,\dots,K_j\}$ and 
\item $ {D}^{(2)}(j,i)=\{I(\m^{(k_i)}_j;\m^{(k_j)}_i | \z_j),  k_i=1,\dots,K_i, k_j=1,\dots,K_j\}$
\item $ {D}^{(3)}(i,j)=\{I(\z_i;\m^{(k_j)}_j ),  k_j=1,\dots,K_j\}$, 
\item $ {D}^{(3)}(j,i)=\{I(\z_j, \m^{(k_i)}_i), k_i=1,\dots,K_i\}$

\end{enumerate}
where $k_i=1,\dots,K_i$, $k_j=1,\dots,K_j$ 

\item creates a vector of descriptors 
\begin{multline}
\label{eq:descr}
x=[ I, \mathcal{Q}(\hat{P}_i), \mathcal{Q}(\hat{P}_j),  \mathcal{Q}(\hat{D}^{(1)}(i,j)), 
\mathcal{Q}(\hat{D}^{(1)}(j,i)),\\
\mathcal{Q}(\hat{D}^{(2)}(i,j)), \mathcal{Q}(\hat{D}^{(2)}(j,i)),
\mathcal{Q}(\hat{D}^{(3)}(i,j)), \mathcal{Q}(\hat{D}^{(3)}(j,i))]
\end{multline}
where $\hat{P}_i$ and $\hat{P}_j$ are the distributions of the populations $\{P_i^{(k_i)} \}$ and $\{P_j^{(k_j)} \}$ , $\hat{D}^{(h)}(i,j)$ denotes the distribution of the corresponding population $D^{(h)}(i,j)$ and
$\mathcal{Q}$ returns a set of sample quantiles of a distribution (in the experiments we set the quantiles to 0.1, 0.25, 0.5, 0.75, 0.9). %

\end{enumerate}
The vector $x$ can be then derived from observational data and used to create a vector
of descriptors to be used as inputs in a supervised learning paradigm.

The rationale of the algorithm is that the asymmetries between $M_i$ and $M_j$ (e.g. Table~\ref{tab:sep})  induce
an asymmetry on the distributions  $\hat{P}$,  and $\hat{D}$  and that the quantiles of those distributions 
provide information about the directionality of causal link ($\z_i \rightarrow \z_j$ or $\z_j \rightarrow \z_i$.) In other terms we
expect that the distribution of these variables should return useful information about which is the cause and the effect. 
Note that these distributions would be more informative if we were able to rank the terms of the Markov Blankets
by prioritizing the direct causes (i.e. the terms $\cc_i$ and $\cc_j$) since these terms play a major role in the asymmetries of Table~\ref{tab:sep}.
The D2C algorithm can then be  improved by choosing an appropriate Markov Blanket selector algorithms, 
like the mIMR filter.

In the experiments (Section~\ref{sec:expe}) we  derive the information
terms as difference between (conditional) entropy terms (see Equations~(\ref{eq:mi}) and~(\ref{eq:cond_mi})) which are themselves estimated by a Lazy Learning regression algorithm~(\cite{Bontempi99BirattariBersini}) by making an assumption of Gaussian noise. Lazy Learning returns a leave-one-out estimation of conditional variance which can be easily transformed in entropy
under the normal assumption~(\cite{Cover06}). The (conditional) mutual information terms are then obtained by
using the relations~(\ref{eq:mi}) and~(\ref{eq:cond_mi}). 

\subsection{Complexity analysis}
As suggested by the reviewers it is interesting to make a complexity analysis of the approach: first it is important
to remark that since the D2C approach relies on a classifier, its learning phase can be time-consuming and dependent on  the 
number of samples and dimension. However, this step is supposed to be performed only once and  from
the user perspective it is more relevant to consider the cost in the testing phase.
Given two nodes for which a test of the existence of a causal link is required, three steps have to
be performed:
\begin{enumerate}
\item computation of the Markov blankets of the two nodes. The information filters we used have a complexity $O(C n^2)$
where $C$ is the cost of the computation
of mutual information~(\cite{meyer2014bontempi}).  In case of very large $n$ this complexity may be bounded by having the 
filter preceded by a ranking algorithm with  complexity  $O(C n)$. Such ranking may limit the number of features taken into consideration  by the filters to $n' < n$ reducing then considerably the cost.
\item once a number $K_i$ ($K_j$) of members of MB$_i$ (MB$_j$)  have been chosen, the rest of the procedure
has a complexity related to the estimation of a number $O(K_i K_j)$ of descriptors. In this paper we used a local learning regression algorithm
to estimate the conditional entropies terms. Given that each regression involves at most three terms, the complexity is essentially related linearly to the number $N$ of samples 
\item the last step consists in the computation of the Random Forest predictions on the test set. Since the RF has been already trained, the complexity 
of this step depends only on the number of trees and not on the dimensionality or number of samples.
\end{enumerate}
For each test, the resulting complexity has then a cost of the order $O(Cn + C n'^2+K_i K_j N)$. It is important to remark that an advantage of D2C is that, if we are interested in predicting the causal relation between two variables only, we are not forced to
infer the entire adjacency matrix (as typically the case in constraint-based methods). 
This mean also that the computation of the entire matrix can be easily made parallel.

\section{Experimental validation}
\label{sec:expe}

In this section the D2C (Section~\ref{sec:D2C}) algorithm is assessed
in a set of synthetic experiments and published datasets.

\subsection{Synthetic data}

This  experimental session addresses the problem of inferring causal
links from synthetic data generated for linear and non-linear DAG configurations of different sizes. All the variables are continuous, and the dependency between children and parents is modelled by the additive relationship
\begin{equation}
\label{eq:DAGdep}
x_{i} =  \sum\limits_{j \in par(i)} f_{i,j}(x_{j}) + \epsilon_{i}, \qquad i=1,\dots,n
\end{equation}
where the noise $\epsilon_{i} \sim N(0,\sigma_{i})$ is Normal, $f_{i,j}(x)   \in  L(x)$
and three sets of continuous functions are considered:
\begin{itemize}
\item {\tt linear}:  $ L(x) = \{f \mid f(x)= a_{0} + a_{1}x \}$
\item {\tt quadratic}:  $L(x) =  \{f \mid f(x)= a_{0} + a_{1}x + a_{2}x^{2} \}$
\item {\tt sigmoid}:  $L(x) =  \{f \mid f(x)= \frac{1}{exp(a_{0} + a_{1}x} \}$
\end{itemize}
In order to assess the accuracy with respect to dimensionality, we considered three network sizes:
\begin{itemize}
\item {\tt small}:  number of nodes $n$ is uniformly sampled in the interval $[20,30]$, 
\item {\tt medium}:  number of nodes $n$ is uniformly sampled in the interval $[100,200]$,
\item {\tt large}:  number of nodes $n$ is uniformly sampled in the interval $[500,1000]$, \\
\end{itemize}
The assessment procedure relies on the generation of a number of DAG structures\footnote{We used the function {\tt random$\_$dag}   from R package gRbase~(\cite{grBase}).} and 
the simulation, for each of them, of  $N$ (uniformly random in  $[100,500]$) node observations
according to the dependency~(\ref{eq:DAGdep}).
In each dataset we removed the observations of five percent of the variables in order to introduce unobserved variables.

For each DAG, on the basis of its structure and the dataset of observations, we collect a number of pairs $\langle x_d,y_d \rangle$, where $x_d$ is the descriptor vector returned by~(\ref{eq:descr}) and $y_d$ is the class denoting the existence (or not) of the causal link in the DAG topology.

The D2C training set is made of $D=6000$ pairs $\langle x_d,y_d \rangle$ and is obtained by generating 750
DAGs and storing for each of them the descriptors
associated to 4 positives examples (i.e. a pair where the node $z_{i}$ is a direct cause of $z_{j}$) and 4 negatives examples (i.e. a pair where the node $z_{i}$ is not a direct cause of $z_{j}$). 
A Random Forest classifier is trained on the balanced dataset: we use the implementation from the R package {\tt randomForest}~(\cite{RF}) with default setting. 

The independent test set is obtained by considering an independent number of
simulated DAGs. We consider 190 DAGs for the small and medium configurations
and 90 for the large configuration. For each testing DAG we select 4 positives examples (i.e. a pair where the node $z_{i}$ is a direct cause of $z_{j}$) and 6 negatives examples (i.e. a pair where the node $z_{i}$ is not a direct cause of $z_{j}$). 
The predictive accuracy of  the trained Random Forest classifier is then assessed on the test set. 

\begin{figure}
\vskip 0.2in
\begin{center}
\includegraphics[width=\columnwidth]{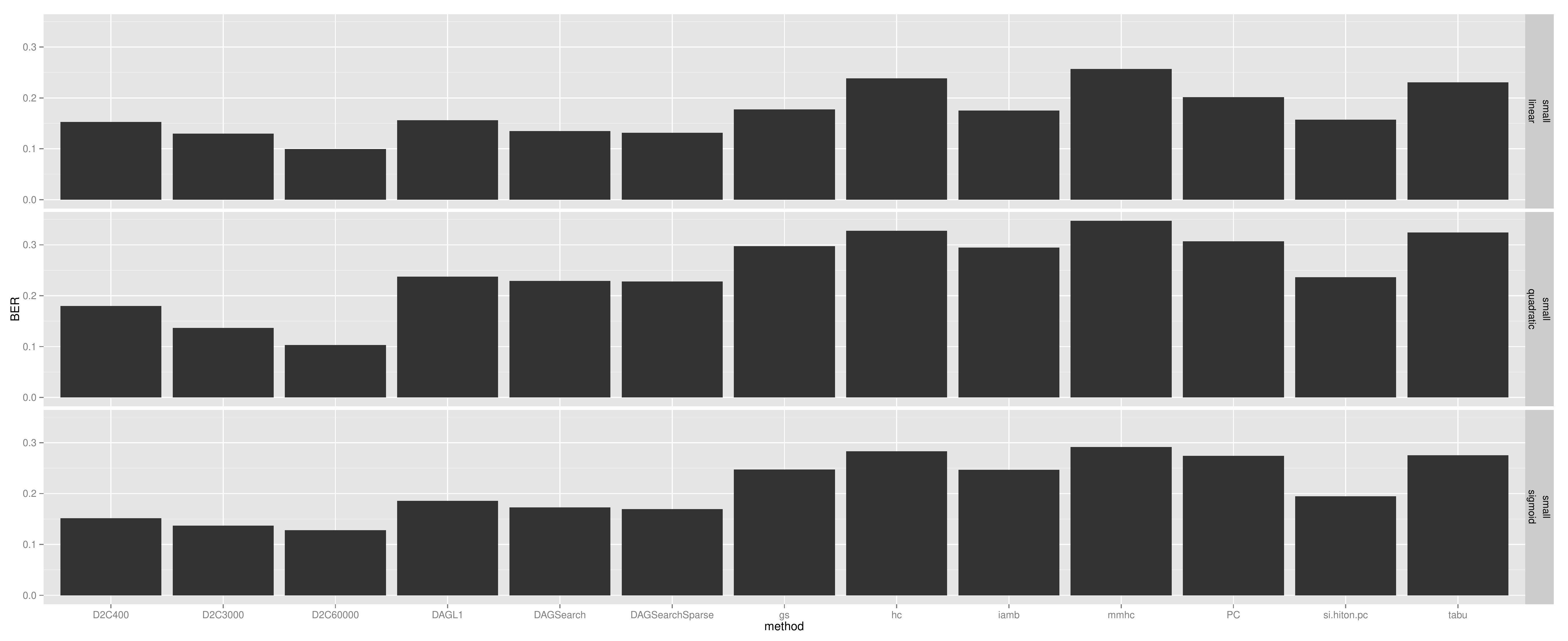}
\caption{Balanced Error Rate of the different methods for small size DAGs and three types of dependency (top: linear, middle: quadratic, bottom: sigmoid). 
The notation D2Cx stands for D2C with
a training set of size $x$.\label{fig:BER.small}}
\end{center}
\vskip -0.2in
\end{figure}

The D2C approach is compared in terms of classification accuracy (Balanced Error Rate (BER)) to
several state-of-the-art approaches implemented and described in the packages {\tt bnlearn}~(\cite{Scutari10}), {\tt pcalg}~(\cite{pcalg}), and {\tt daglearn}
\footnote{http://www.cs.ubc.ca/$\sim$murphyk/Software/DAGlearn/}: 
\begin{itemize}
\item  {\tt DAGL1}: DAG-Search score-based algorithm with potential parents  selected with a L1 penalization~(\cite{Schmidt07etal}).   
\item  {\tt DAGSearch}: unrestricted DAG-Search score-based algorithm (multiple restart greedy hill-climbing, using edge additions, deletions, and reversals)~(\cite{Schmidt07etal}),
\item  {\tt DAGSearchSparse}:  DAG-Search score-based algorithm with potential parents  restricted to the $10$ most correlated features~(\cite{Schmidt07etal}), 
\item {\tt gs}: Grow-Shrink constraint-based structure learning algorithm,
\item  {\tt hc}: hill-climbing score-based structure learning algorithm,
\item {\tt iamb}: incremental association MB constraint-based structure learning algorithm,
\item  {\tt mmhc}: max-min hill climbing hybrid structure learning algorithms,
\item  {\tt PC}: Estimate the equivalence class of a DAG using the PC algorithm (this method was used only for the small size configuration (Figure~\ref{fig:BER.small}) 
for computational time reasons)
\item {\tt si.hiton.pc}: Semi-Interleaved HITON-PC local discovery structure learning algorithms,
\item {\tt tabu}: tabu search score-based structure learning algorithm,
\end{itemize}

\begin{figure}
\vskip 0.2in
\begin{center}
\includegraphics[width=\columnwidth]{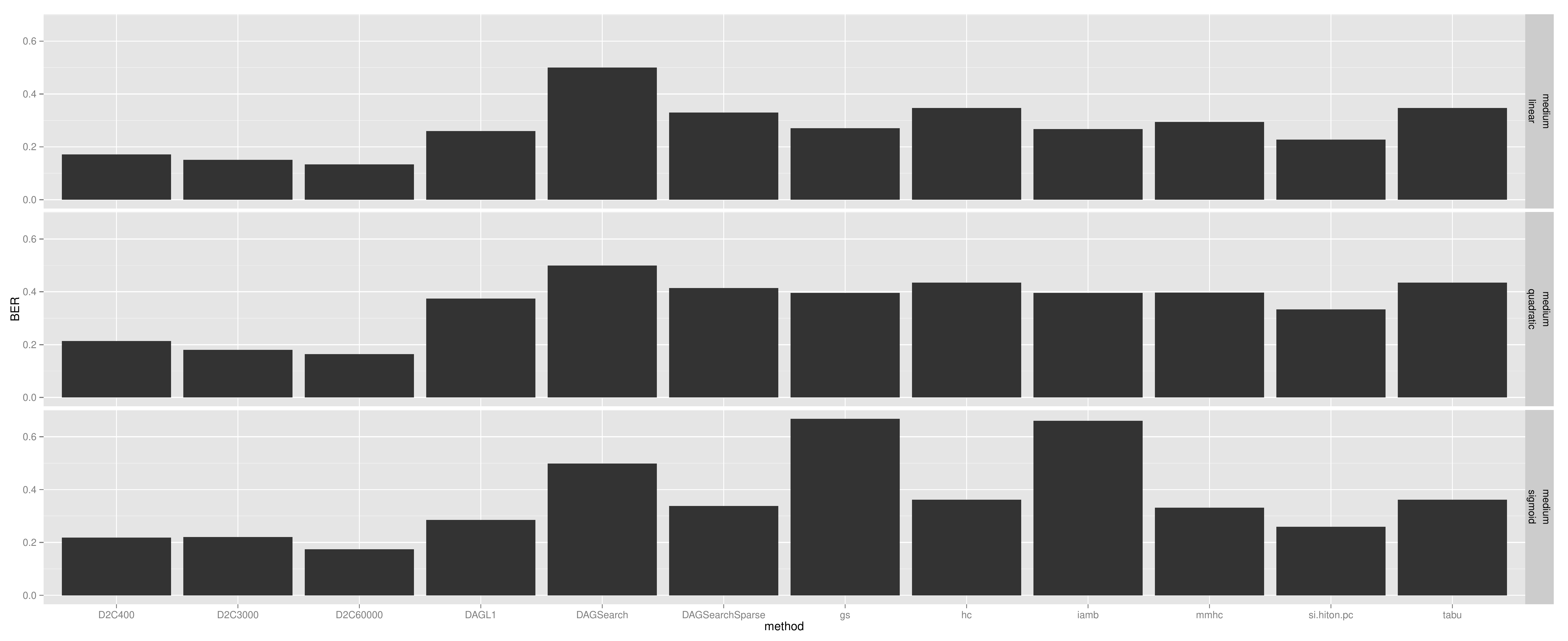}
\caption{Balanced Error Rate of the different methods for medium size DAGs and three types of dependency (top: linear, middle: quadratic, bottom: sigmoid). The notation D2Cx stands for D2C with
a training set of size $x$.\label{fig:BER.medium}}
\end{center}
\vskip -0.2in
\end{figure}

\begin{figure}
\vskip 0.2in
\begin{center}
\includegraphics[width=\columnwidth]{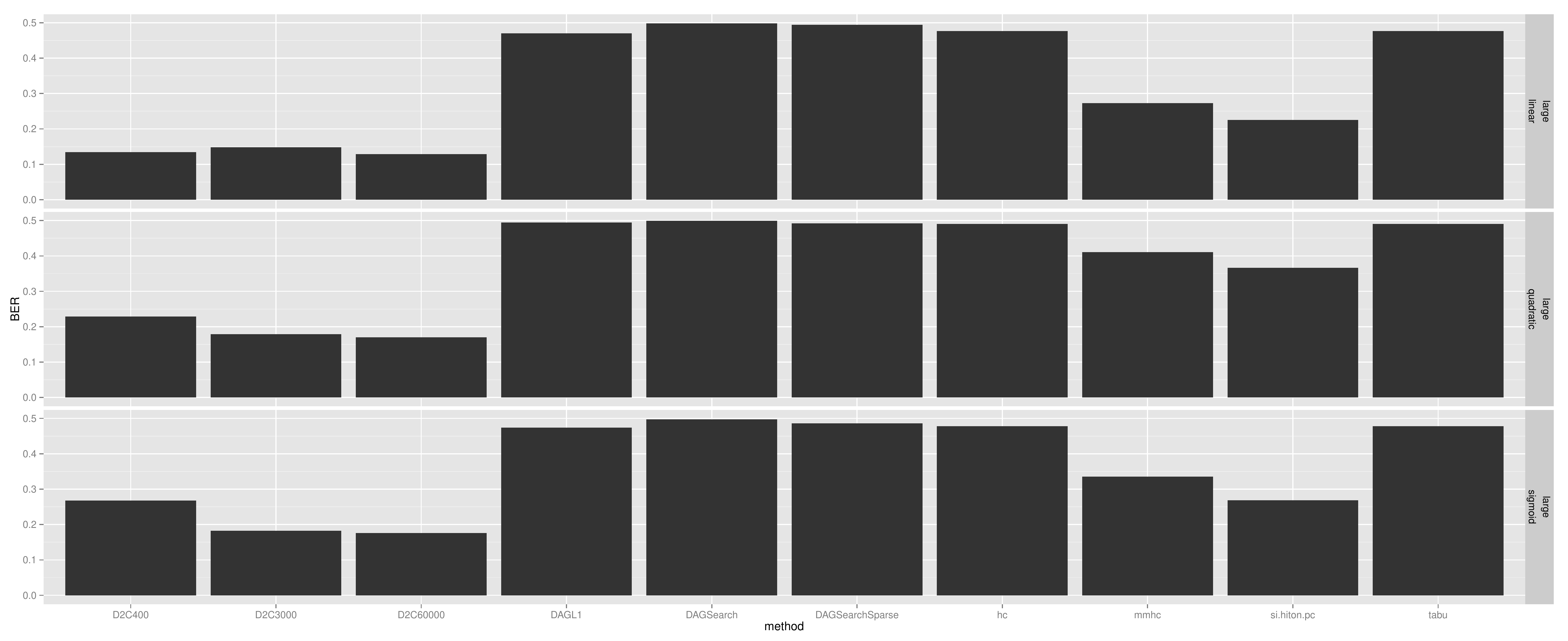}
\caption{Balanced Error Rate of the different methods for large size DAGs and three types of dependency (top: linear, middle: quadratic, bottom: sigmoid). The notation D2Cx stands for D2C with
a training set of size $x$.\label{fig:BER.large}}
\end{center}
\vskip -0.2in
\end{figure}

The BER of three versions of the D2C method (training set size equal to 400, 3000 and 6000 respectively) are compared
to the BER of  state-of-the-art methods in Figures~\ref{fig:BER.small} (small), Figure~\ref{fig:BER.medium} (medium)  and Figure~\ref{fig:BER.large} (large).
Each subfigure corresponds to the three types of stochastic dependency (top: linear, middle: quadratic, bottom: sigmoid).

A series of considerations can be made on the basis of the experimental results:
\begin{itemize}
\item the n-variate approach D2C obtains competitive results with respect to several state-of-the-art techniques  in the linear case,
\item the improvement of D2C wrt state-of-the-art techniques (often based on linear assumptions) tends to increase when we move to more nonlinear configurations,
\item the accuracy of the D2C approach improves by increasing the number of training examples,
\item with a small number of examples (i.e. $N=400$) it is already possible to learn a classifier D2C
whose accuracy is competitive with state-of-the-art methods. 
\end{itemize}

The D2C code is available in the CRAN R package {\tt D2C}~\cite{D2Cpackage}.

\subsection{Published data}
The second part of the assessment relies on the simulated and resimulated datasets proposed in~(\cite{Aliferisetal2010}, Table 11).
These 103 datasets were obtained by simulating data from known Bayesian networks and also by resimulation, where real data is used to elicit a causal network and then data is simulated from the obtained network.
We split the 103 datasets in two portions: a training portion (made of 52 sets) and a second portion (made of 51 sets)  for testing.
This was done in order to assess the accuracy of two versions of the D2C algorithm: the first uses as training set
only the samples generated in the previous section, the second includes in the training set also the 52 datasets of the training portion.
The goal is to assess the generalization accuracy of the D2C algorithm with respect to DAG distributions never encountered before and 
not included in the training set.
In this section we compare D2C to a set of algorithms implemented by the \emph{Causal Explorer} software~(\cite{tsamardinos03aliferissoftware})\footnote{Note that we use \emph{Causal Explorer} here because, unlike {\tt bnlearn} which estimates the entire adjacency matrix,  it returns a ranking of the inferred causes for a given node.}:
\begin{itemize}
\item {\tt GS}: Grow/Shrink algorithm
\item {\tt IAMB}: Incremental Association-Based Markov Blanket
\item {\tt IAMBnPC}: IAMB with PC algorithm in the pruning phase
\item {\tt interIAMBnPC}: IAMB with PC algorithm in the interleaved pruning phase
\end{itemize}
and two filters based on information theory, mRMR~(\cite{Peng05LongDing}) and mIMR~(\cite{bontempi2010causal}).
The comparison is done as follows: for each dataset and for each node (having at least a parent)  the causal inference techniques 
return the ranking of the inferred parents. The ranking is assessed  in terms of the average of Area Under the Precision Recall Curve (AUPRC) 
and a t-test is used to assess if the set of AUPRC values is significantly different between two methods. Note that the higher the AUPRC the
more accurate is the inference method.

The summary of the paired comparisons is reported in Table~\ref{table:AUPRCsynth} for the D2C algorithm trained on the synthetic data only
and in Table~\ref{table:AUPRCtrained} for the D2C algorithm trained on both synthetic data and the 52 training datasets.

\begin{table}
\hspace{-1cm}
\begin{tabular}{c||c|c|c|c|c|c}
			& GS 		& IAMB 		& IAMBnPC 	& interIAMBnPC 	& mRMR 		& mIMR \\
\hline
W-L 	& 48-3 (32-0)  & 43-8 (21-0) 	& 46-5 (26-0) 	& 46-5 (25-0) 		& 42-9 (17-0) 	& 34-17 (12-0)  \\
\end{tabular}
\caption{D2C trained on synthetic data only: number of datasets for which D2C  has an AUPRC (significantly (pval $<0.05$)) higher/lower 
than the method in the column. W-L stands for Wins-Losses.\label{table:AUPRCsynth}}
\end{table}

\begin{table}
\hspace{-1cm}
\begin{tabular}{c||c|c|c|c|c|c}
			& GS 		& IAMB 		& IAMBnPC 	& interIAMBnPC 	& mRMR 		& mIMR \\
\hline
W-L 	& 49-2 (36-0)  & 49-2 (27-0) 	& 49-2 (32-0) 	& 49-2 (32-0) 		& 42-9 (17-0) 	& 46-5 (19-1)  \\
\end{tabular}
\caption{D2C  trained on synthetic data and 52 training datasets: number of datasets for which the D2C has an AUPRC  (significantly (pval $<0.05$)) higher/lower  
than the method in the column. W-L stands for Wins-Losses.\label{table:AUPRCtrained}}
\end{table}

It is worthy to remark that
\begin{itemize}
\item the D2C algorithm is extremely competitive and outperforms the other techniques taken into consideration,
\item the D2C algorithm is able to generalize to DAG with different number of nodes and different distributions also when trained only
on synthetic data simulated on linear DAGs,
\item the D2C algorithm takes advantage from the availability of more training data and in particular of training data related to the causal
inference task of interest, as shown by the improvement of the accuracy from Table~\ref{table:AUPRCsynth} to Table~\ref{table:AUPRCtrained},
\item  the two filters (mRMR and mIMR) algorithm appears to be the least inaccurate among the state-of-the-art algorithms,
\item though the D2C is initialized with the results returned by the mIMR algorithm, it is able to improve its output and to significantly outperform it.
\end{itemize}

\section{Conclusion}
Two attitudes are common with respect to causal inference
for observational data.
The first is  pessimistic  and motivated by the consideration that \emph{correlation (or dependency) does not imply causation}.
The second is  optimistic  and driven by the fact that \emph{causation implies correlation (or dependency)}.
This paper  belongs evidently to the second school of thought and relies on the confidence that causality leaves footprints in the form of stochastic dependency and that these footprints can be detected to retrieve causality from observational data. 
The results of the ChaLearn challenge and the preliminary results of this paper confirm the potential of machine learning approaches in predicting the existence of causality links on the basis of statistical descriptors of the dependency. We are convinced that this will open a new research direction where 
learning techniques may be used to reduce the degree of uncertainty about the existence of a causal relationships also in indistinguishable configurations
which are typically not addressed by conditional independence approaches.

Further work will focus on 1) discovering additional features of multivariate
distributions to improve the accuracy 2) addressing and assessing other related classification problems (e.g. predicting if a variable is an ancestor 
or descendant of a given one) 3) extending the work to partial ancestral graphs (e.g. exploiting the logical relations presented in~(\cite{Claassen11Heskes})) 4 ) extending the validation to real datasets and configurations with a still larger number of variables (e.g. network inference
in bioinformatics).  

\bibliographystyle{plain}
\bibliography{dep2caus}

\end{document}